\pdfoutput=1

\documentclass[11pt]{article}

\usepackage{acl}

\usepackage{times}
\usepackage{latexsym}
\usepackage{multirow}
\usepackage{ulem}
\usepackage[T1]{fontenc}

\usepackage[utf8]{inputenc}

\usepackage{enumitem}
\usepackage[utf8]{inputenc}
\DeclareUnicodeCharacter{FF0C}{,}

\usepackage{microtype}

\usepackage{inconsolata}
\usepackage{amsmath}
\usepackage{graphicx}
\usepackage{subfigure}

\title{Enhancing the Preference Extractor in Multi-turn Dialogues: From Annotating Disasters to Accurate Preference Extraction}

\author{
    \textbf{Cheng Wang\textsuperscript{1,}\textsuperscript{\dag}},
    \textbf{Ziru Liu\textsuperscript{2,}\textsuperscript{\dag}},
    \textbf{Pengcheng Tang\textsuperscript{1}\textsuperscript{\dag}},
    \textbf{Mingyu Zhang\textsuperscript{1}\textsuperscript{\dag}},
    \textbf{Quanyu Dai\textsuperscript{2}},
    \textbf{Yue Zhu\textsuperscript{1,}\textsuperscript{\ddag}}
    \\
    \textsuperscript{1}Huawei Technologies Co., Ltd.
    \textsuperscript{2}Huawei Noah's Ark Lab
    \\
    \{wangcheng250, liuziru6, tangpengcheng12, zhangmingyu14, daiquanyu, zhuyue9\}@huawei.com
}

\begin{document}
\maketitle

\footnotetext[1]{$^{\dag}$These authors contributed equally to this work.}
\footnotetext[2]{$^{\ddag}$Corresponding author.}

\begin{abstract}

Identifying user preferences in dialogue systems is a pivotal aspect of providing satisfying services. Current research shows that using large language models (LLMs) to fine-tune a task-specific preference extractor yields excellent results in terms of accuracy and generalization. However, the primary challenge stems from the inherent difficulty in obtaining high-quality labeled multi-turn dialogue data. Accurately tracking user preference transitions across turns not only demands intensive domain expertise and contextual consistency maintenance for annotators (termed \textbf{``Annotating Disaster''}) but also complicates model training due to error propagation in sequential dependency learning. Inspired by the observation that multi-turn preference extraction can be decomposed into iterative executions of one-turn extraction processes. We propose a novel dialogue data generation framework named \textbf{IterChat}. First, we construct a new data format that categorizes the dialogue data into attributed historical preferences and one-turn dialogues. This reduces the probability of annotation errors and improves annotation efficiency. Then, to generate a high-quality and diverse dialogue dataset, we adopt GPT4 to pre-define the preference slots in the target preference extractor task and then randomly sample the subset of the slots and their corresponding schema values to create the dialogue datasets. Experimental results indicate that fine-tuning or only few-shot prompting with the new dialogue format yields superior performance compared to the original multi-turn dialogues. Additionally, the new data format improves annotator efficiency with a win rate of 28.4\% higher than the original multi-turn dialogues.
\end{abstract}

\section{Introduction}

A significant challenge in web-based customer support lies in the efficient recognition of user preferences within service dialogues \cite{malik2024pearl,cheng2021exploring,shin2022dialogue}. Unlike traditional search-based services that process single-shot queries, multi-turn conversations necessitate the identification of dynamically evolving user preferences embedded within the dialogue \cite{pai2024survey,han2023information,feng2021sequence}. Recent studies adopt the Large Language model (LLM) to empower the ability to accurately track user preferences in real-time multi-turn user-system dialogues, thereby enabling the provision of tailored services \cite{xu2024chain,guo2022beyond,ravuru2022multi}. In contrast to the entity extraction task, which focuses on identifying and classifying specific entities within the text, preference extraction involves analyzing and deriving users' emotions, interests, and intentions from the text, requiring a deeper level of comprehension \cite{yi2024survey,feng2024tasl}. This capability can substantially enhance both the customer experience and the quality of service, while simultaneously supporting business intelligence initiatives for companies \cite{zhou2022dialogue,qixiang2022exploiting}.

Recent LLM-based preference extraction focused on leveraging prompt engineering combined with few-shot examples \cite{feng2023towards,xu2024chain,malik2024pearl}. These methods utilize prompts to assign specific roles to LLMs and define the slots to be extracted. However, the few-shot performance of leading LLM, such as GPT-4, still falls short of the state-of-the-art supervised methods \cite{qi2023preserving}, especially when user queries are broad, ambiguous, and upper funnel \cite{kim2024meganno+,heck2023chatgpt}. Hence, some works start to utilize the fine-tuning technique to train the foundation model with the open source datasets \cite{feng2023towards}. However, practical commercial services, such as e-commerce, require a high level of accuracy in identifying complex user preference slots and require customizing additional slots to meet personalized services \cite{malik2024pearl}, as this directly impacts the ability to provide users with suitable and satisfactory products.

Therefore, creating a high-quality customized dialogue dataset for a task-oriented domain is crucial to developing a well-performing preference extractor \cite{li2023evaluating}. However, even for experts, tracing the preference transition and annotating an accurate label for the multi-turn conversation is challenging. This is because preference extraction in dialogue data not only requires attention to non-standardized and ambiguous user utterances but also involves continuously adding, removing, or updating preferences based on the user's reactions to system responses. Consequently, acquiring a large-scale golden dataset to train a task-oriented preference extractor is costly and inefficient, a phenomenon we refer to as the \textbf{``Annotating Disaster''}. For more details about the annotating disaster, please refer to Section \ref{sec:Multi-Turn Dialogues and Preference Extraction} and Figure \ref{fig:p1}. Another significant challenge is that long conversational contexts make model training more difficult, as the cumulative errors in the preference extraction steps tend to accumulate as the dialogue context grows.

To address the aforementioned challenges, we propose a novel dialogue data generation framework named \textbf{IterChat}, which is designed to be both annotation-friendly and training-efficient. The framework is inspired by the observation that \textit{multi-turn preference extraction can be decomposed into iterative executions of one-turn extraction processes.} This insight implies that modeling preference evolution through atomic single-turn operations can reduce annotation complexity and minimize error propagation during model training.  Specifically, we transform the traditional multi-turn dialogue data into a new data format, which categorizes the dialogue data into historical preferences and the most recent one-turn dialogues. For annotators, the refined dialogue format enables them to annotate the preference transition only once. For fine-tuning LLMs, this new data format does not require long context as input, thereby saving tokens and allowing the model to learn extraction rules from simpler input. Additionally, to overcome the limitation of systematic biases inherent in LLMs and the diversity of the generated dialogue data, we utilize the assistance of LLMs to define the preference slots that need to be extracted for task-oriented preference extractors. We then randomly sample slots and their state values to generate the new form of dialogue datasets.

The main contributions of our work are summarized as follows.
\begin{itemize}[leftmargin=*]
    \item  We transform the traditional multi-turn dialogue data into a new data format that categorizes dialogues into historical preferences and the most recent one-turn dialogues. This refined format reduces annotation errors improves efficiency for annotators, and optimizes the fine-tuning process by simplifying input for LLMs, thus saving tokens and enhancing the learning process.

\item We propose a method to overcome the limitations of systematic biases in LLMs and the diversity of generated dialogue data by utilizing LLMs to define task-oriented preference slots. These slots are randomly sampled along with their state values to generate new dialogues, facilitating the development of accurate preference extractors.

\item Experimental results demonstrate that fine-tuning or few-shot prompting with the new dialogue format yields superior performance compared to the original multi-turn dialogues. Moreover, this new data format enhances annotator efficiency, achieving a 28.4\% higher win rate than the original multi-turn dialogues.
\end{itemize}

\section{Related Works}
\subsection{Preference Extraction on LLM-based Multi-turn Dialogue}
Preference extraction, also known as Dialogue State Tracking (DST), aims to track hidden preferences embedded in conversations to fulfill user goals in task-oriented dialogue systems \cite{gu2024plan,gu2024two}. With the emergence of LLMs exhibiting remarkable zero-shot capabilities, researchers have begun to explore using LLMs as task-oriented preference extractors. For instance, both \cite{lee2021dialogue} and \cite{yang2023dual} proposed a prompt-tuning method that leverages domain-specific prompts and contextual information to improve the performance of the preference extraction task. \cite{xu2024chain} constructed chain-of-thought reasoning for the preference extraction task by extracting multiple system-user utterance pairs from dialogue history that alter slot values. \cite{malik2024pearl} proposed a framework in which LLMs first summarize user preferences from dialogues, followed by a dynamic example retrieval module that stores and retrieves ICL examples. Recent \cite{feng2023towards,an2023dialogue,moghe2021cross} studies have found that few-shot learning performance remains inadequate. Consequently, research has shifted towards fine-tuning techniques to develop more effective preference extractors. Although various methods focus on the preference extraction task, obtaining large amounts of high-quality task-oriented labeled dialogue data to address complex real-world dialogue scenarios remains a challenge. This is because annotators often face difficulties in annotating multiple turns of slot-value pairs, which can be time-consuming and complex.

\subsection{Labeled Dialogue Data for Preference Extraction}
Some researchers have contributed a series of labeled datasets for the preference extraction task. For example, MultiWOZ 2.2 \cite{eric2019multiwoz} is a multi-domain task-oriented dialogue dataset that includes more than 10,000 dialogues that span 8 domains. Additionally, the Schema-Guided Dialogue (SGD) \cite{rastogi2020towards} dataset includes over 16,000 conversations between users and virtual assistants, which encompass 26 services in 16 domains, such as events, restaurants, and media. However, with the increasing number of online services providing dialogue interfaces, current open-source datasets struggle to cover all specific scenarios. Moreover, the preference slots available in these open-source datasets are limited, making it challenging for service providers to build an accurate preference extractor to handle varied and changing user preferences. Therefore, we propose a dialogue data generation framework named IterChat to help service providers quickly construct labeled dialogue datasets for their own domains.

\section{Preliminaries}
This section provides an overview of LLM-based multi-turn dialogue systems and the associated challenges of multi-turn preference extraction.

\subsection{Multi-Turn Dialogues and Preference Extraction}
\label{sec:Multi-Turn Dialogues and Preference Extraction}
In task-oriented dialogue systems, users typically engage in multi-turn interactions with a chatbot to iteratively clarify, adjust, or refine their preferences \cite{feng2023towards}. This process can be modeled as a sequence of dialogue pairs: $\{(Q_1, A_1), (Q_2, A_2), \dots, (Q_T, A_T)\},$ where $Q$ represents the user’s input queries, and $A$ represents the chatbot's responses. 
Each pair represents a single dialogue turn. The dialogue context at turn $t$ includes the entire history of interactions up to that point, incorporating both the user’s queries and the chatbot’s responses, and is denoted as: $X_t = \{(Q_1, A_1), (Q_2, A_2), \dots, (Q_t, A_t)\}$. This context plays a critical role in understanding the evolving preferences and intentions of the user.

The primary task in preference extraction for dialogue systems is to identify key pieces of information from a conversation that reflect the user's current preferences \cite{malik2024pearl}. These preferences are typically expressed through various slots, each representing a specific aspect of the user’s intent or requirement. At any given turn \(t\), the dialogue information can be represented as a set of preference slots, each associated with a particular entity value, denoted as:
$Y_t = \{(P_1: \textbf{V}_{1,t}), (P_2: \textbf{V}_{2,t}), \dots, (P_N: \textbf{V}_{N,t})\}$, where \(P_i\) is the preference slot, and \(\textbf{V}_{i,t}\) is the corresponding values of that slot at turn \(t\). For instance, in an e-commerce scenario, a user might express their preference for a product in terms of a slot such as ‘‘$\left\langle price\right\rangle$", which indicates the expected price range. The corresponding value for this slot could be something like ‘‘less than \$50", which specifies the user's preference in more detail. Other common preference slots in such scenarios could include ‘‘$\left\langle color\right\rangle$", ‘‘$\left\langle brand\right\rangle$", or ‘‘$\left\langle size\right\rangle$", each reflecting a specific dimension of the user's choice.

In multi-turn dialogues, large language models (LLMs) are commonly used to extract preference slots from a sequence of question-answer pairs. However, existing LLMs face inherent limitations in retaining long-term memory across extended conversations. This often leads to a phenomenon known as "preference slot oblivion", where the model loses track of earlier preferences as the dialogue progresses, resulting in inconsistencies in its understanding. To address this challenge in preference extraction within multi-turn dialogues, we propose a novel approach in the next section that reorganizes the problem into an incremental preference evolution framework. In this framework, the learning objective for the LLM is to first extract preference slots and values from the most recent one-turn dialogue. Then, it combines the user’s historical preferences with the latest preferences from the current dialogue turn to form the most up-to-date user preference.

\section{The Proposed Framework}
\label{sec:method}

\begin{figure*}
    \centering    \includegraphics[width=0.98\linewidth]{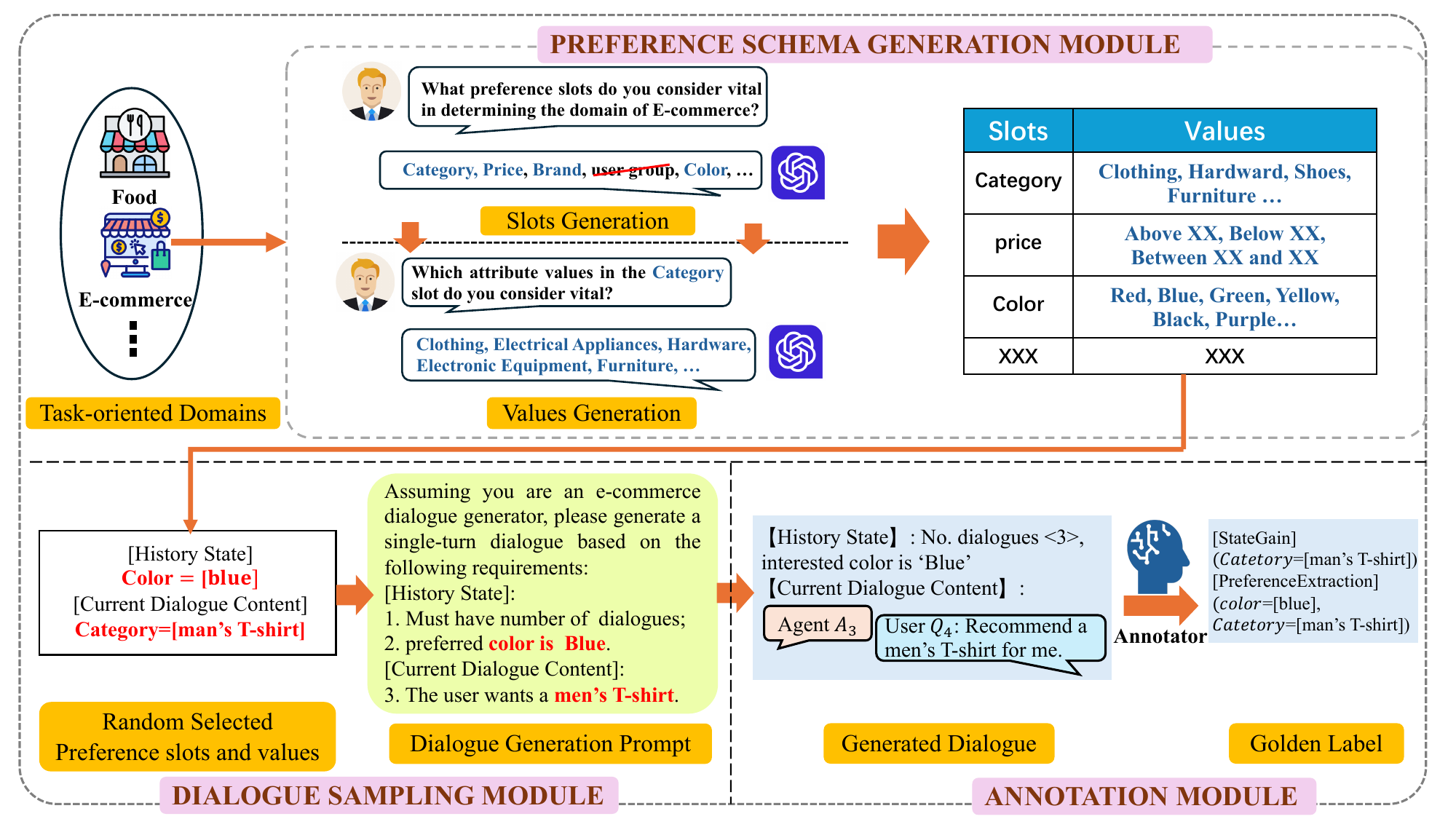}
    \caption{Overview Framework of IterChat Data Generation}
    \label{fig:framework}
\end{figure*}

In this section, we provide a detailed explanation of our proposed \textbf{IterChat} and corresponding data format, along with the annotation process and the overall framework for IterChat data generation. This includes an in-depth description of each module involved in the pipeline. As illustrated in Figure \ref{fig:framework}, the main framework consists of four key modules: the preference schema module, the dialogue sampling module, the annotation module, and the agent tuning module.

\begin{figure} [h]
    \centering    \includegraphics[width=0.98\linewidth]{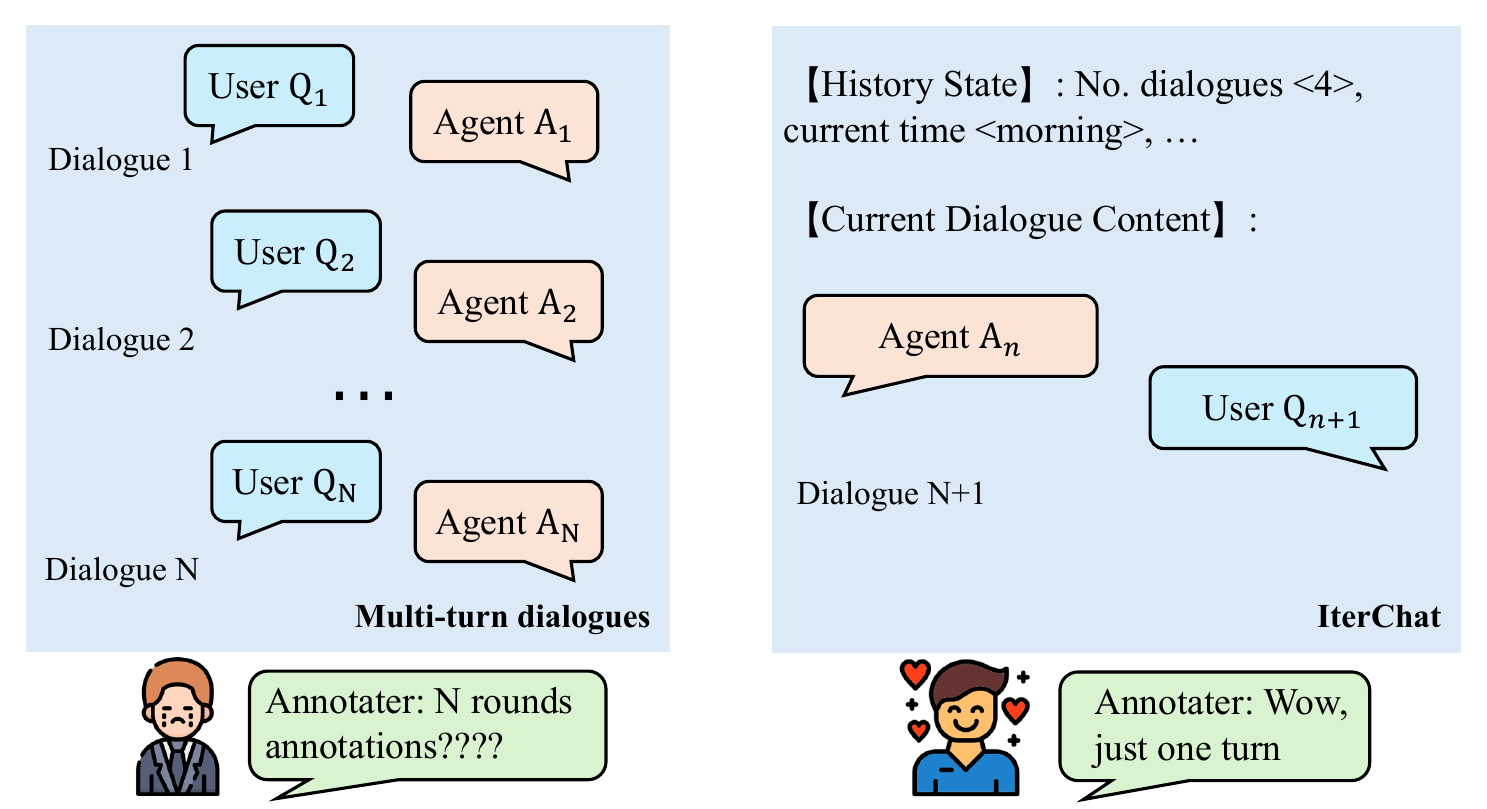}
    \caption{Comparison of multi-turn dialogues and IterChat data}
    \label{fig:p1}
\end{figure}

\subsection{Incremental Preference Evolution}

Given a dialogue sequence up to the \((t+1)\)-th turn: $X_{t+1} = \{(Q_1, A_1), \dots, (Q_{t+1}, A_{t+1})\}$, instead of directly extracting the preference \(Y_{t+1}\), we first summarize the preference information from the previous \(t\)-turn dialogue, \(X_t\), to obtain the current preference \(Y_t\). Subsequently, we extract the preference from the most recent one-turn dialogue \((A_{t}, Q_{t+1})\) based on the context \(Y_t\), yielding the preference gain \(G_{t+1}\), which involves updating the preference slots. Finally, we combine the historical preference \(Y_t\) with the newly extracted preference gain \(G_{t+1}\) to update the current preference, \(Y_{t+1}\).

This iterative framework ensures that, by leveraging both historical preference \(Y_t\) and the most recent dialogue turn \((A_{t}, Q_{t+1})\), we can effectively extract the preference gain \(G_{t+1}\) and obtain the updated preference \(Y_{t+1}\). This methodology effectively prevents preference slot oblivion, offering a more structured and coherent process for maintaining preference consistency throughout the dialogue. Specifically, the learning objective for the LLM is to extract preference slots and values from the most recent one-turn dialogue, then combine the user’s historical preferences (as captured in the History Preference) with the latest preferences from the current dialogue turn to form the most up-to-date user preference. This approach mitigates the problem of preference slot oblivion and ensures that the model can continuously track evolving user needs. Based on this problem definition, we further propose reorganizing multi-turn dialogue data into a new, more efficient format, which will be explained in detail in the next section.

\subsection{An Annotate-friendly data format}
\label{sec:IterChat}
We reorganize multi-turn dialogue data into a new, more efficient format which consists of two main components: 
\begin{itemize} [leftmargin=*]
    \item \textbf{History Preference}: It summarizes the user’s preferences over the previous $n$ turns of dialogue, capturing the evolving context and the user's changing preferences.
    \item \textbf{Most Recent One-Turn Dialogue}: It contains the latest user query and chatbot response, reflecting the immediate context of the ongoing conversation.
\end{itemize}

A comparison between multi-turn dialogues and the IterChat format is illustrated in Figure \ref{fig:p1}. By adopting this structure, we transform the original multi-turn preference extraction problem into a more manageable incremental preference evolution problem.

In addition, we introduce two new annotation outputs for preference extraction in the IterChat format: ‘‘StateGain" and ‘‘PreferenceExtraction". ‘‘StateGain" represents the information gained from the most recent dialogue turn, highlighting the new insights added to the user's preferences. On the other hand, ‘‘PreferenceExtraction" reflects the final set of preference slots after processing user history preference and the latest dialogue preference, representing the chatbot’s understanding of the user’s current preferences. By using this approach, human annotators only need to annotate the preference slots for the most recent one-turn dialogue, significantly reducing the likelihood of annotation errors and improving efficiency.

\subsection{Preference Schema Module}
Effectively extracting user preferences involves monitoring the user’s shifting goals and the system's responses throughout the dialogues. To maintain consistency in understanding user preferences, it is essential to produce structured outputs. This can be achieved by extracting predefined slot-value pairs from the dialogue context at each turn, ensuring that the chatbot can interpret and act on the user’s preferences with clarity and precision.

In the Preference Schema Generation Module, we begin by using GPT‑4 to draft the core preference slots for the target extraction task—the factors that most strongly drive a user’s decisions, such as budget constraints in an e-commerce scenario or preferred location in a travel planning task. We then iteratively refine these into a formal, task‑oriented schema: a concise set of slots, each capturing a distinct facet of preference that must be tracked during dialogue. In an e‑commerce scenario, for instance, the schema might include ‘‘$\left\langle price\right\rangle$", ‘‘$\left\langle brand\right\rangle$", or ‘‘$\left\langle color\right\rangle$". Each slot may have a range of possible values, depending on the user's preferences, GPT‑4o is further applied to determine their corresponding schema value for each preference slots. For example, the ‘‘$\left\langle price\right\rangle$" slot could have values like ‘‘less than \$50", ‘‘between \$100 and \$200", or "None". By structuring the preferences in this way, the dialogue system can consistently track and update user preferences across multiple turns, ensuring that the system’s responses remain aligned with the user's evolving needs.


\subsection{Dialogue Sampling Module}

One of the key advantages of the preference schema is that it enables the generation of high-quality user dialogues. In this section, we outline how we leverage the preference schema to construct IterChat data which consists of the user's history state and the most recent one-turn dialogue, with updates to preference slots that can be tailored to our needs. 

The dialogue sampling process begins with the construction of the ‘‘history state", which includes details such as the number of past dialogues, the current time, and other relevant context information that reflects the conversation’s progression. The ‘‘history state" is generated by randomly sampling detailed contextual information using a Context Agent. This agent is responsible for selecting a variety of factors that summarize the history of the conversation, ensuring the generated state is diverse and representative of different conversational scenarios. Additionally, the Context Agent is tasked with sampling the current state, which includes preference slots of particular interest. These preference slots could correspond to factors like price, color, or brand in an e-commerce scenario, or location and date in a travel planning context.

With both the history state and current state in hand, the Context Agent generates the most recent one-turn dialogue, reflecting the update of preference slots. For instance, if the history state is: 

\[
\text{(‘price' = [‘less than \$50'])}
\]

the current state is:

\[
\text{(‘price' = [‘less than \$50'], ‘color' = [‘red'])}
\]

the corresponding one-turn dialogue could be:

\[
\text{user: ‘‘I like red."}
\]

This approach allows for a smooth transition between states and generates natural dialogues that are contextually relevant and reflective of evolving user preferences. By following this pipeline, we can efficiently sample large quantities of IterChat data, each containing a rich set of preference slots that are of interest. This approach not only supports the generation of diverse dialogues but also ensures that each dialogue remains relevant to the user's preferences. Please refer to Appendix \ref{app: prompt} for the prompt.

\subsection{Annotation Module}
In addition to this structure, we propose a new annotation output for preference extraction in the IterChat format, defined by two components:

\begin{itemize} [leftmargin=*]
    \item \textbf{StateGain:} This represents the information gained from the most recent dialogue. Specifically, it quantifies how much new information has been added to the user’s preference profile after processing the latest interaction. The StateGain helps identify whether the most recent dialogue has refined or introduced new preferences.
    \item \textbf{PreferenceExtraction:} This denotes the final extraction of preference slots based on the dialogue so far, encompassing both the historical context and the latest one-turn. It represents the chatbot’s understanding of the user’s current preferences after incorporating the entire dialogue history and the most recent interaction. The PreferenceExtraction result is a comprehensive set of preference slots, each with an associated value, reflecting the user's intentions. 
\end{itemize}
Note that this process involves more than just expanding or summing up slots. In real-world applications, it requires adhering to multiple inheritance rules to ensure the consistency and accuracy of preference updates. By using the IterChat data format and its corresponding annotations, we can generate diverse and high-quality raw data while enhancing annotator efficiency, as only one-turn annotations are required, compared to the multiple-turn annotations typically needed for full dialogues.

Once the annotated IterChat data is collected and stored in our database, it is used to further fine-tune our dialogue agent in a supervised manner, thereby enhancing its ability to understand and respond to user preferences. The IterChat data format plays a crucial role in optimizing this fine-tuning process by providing a simplified, structured input for large language models (LLMs). By leveraging the concise, one-turn structure of the IterChat format, we significantly reduce the computational overhead, allowing for more efficient training. Moreover, this format ensures that the agent can better capture evolving preferences by maintaining a clear, consistent representation of user goals and system actions. In the next section, we demonstrate that fine-tuning or few-shot prompting with IterChat yields superior performance compared to the original multi-turn dialogues.

\begin{table*}[h]

    \centering
    \resizebox{\linewidth}{!}{
    \begin{tabular}{c|c|c|c|c|c|c|c|c|c|c|c|c|c|c}
    \hline
    \multirow{2}{*}{} & \multirow{2}{*}{Setup} & \multirow{2}{*}{Paradigm} & \multicolumn{4}{|c|}{Foodie} & \multicolumn{4}{|c|}{MultiWOZ-H} & \multicolumn{4}{|c}{MultiDoGO-S} \\
    \cline{4-15}
    & & & EM $\uparrow$ & F1 $\uparrow$ & FED $\downarrow$ &BLEU $\uparrow$ & EM $\uparrow$ & F1 $\uparrow$ & FED $\downarrow$ &BLEU $\uparrow$ & EM $\uparrow$ & F1 $\uparrow$ & FED $\downarrow$ &BLEU $\uparrow$\\
    \hline
    \multirow{4}{*}{few-shot prompting} & PERAL-GPT4& ICL@2 multi-turn & 0.32 &0.8158 & 1.638 & 0.5661  & \textbf{0.5674} & \textbf{0.7863} & \textbf{1.081} &\textbf{0.6217} &0.0833	&0.3919	&3.1636 & 0.3173 \\
 
    & PERAL-GPT4& ICL@2 IterChat & \textbf{0.501} & \textbf{0.9021} & \textbf{0.8157} & \textbf{0.7185} & 0.5379 & 0.7244 & 1.282 & 0.6185 &\textbf{0.1333}	&\textbf{0.518}	&\textbf{2.3565} & \textbf{0.4022} \\
    \cline{2-15}
    & NL2API-GPT4 & ICL@2 multi-turn & 0.3333 & 0.806 & 1.233 & 0.5345 & \textbf{0.5538} & \textbf{0.7399} & \textbf{1.085} & \textbf{0.6203} &0.1166	&0.6204	&3.1351 & 0.4695 \\
    
    & NL2API-GPT4 & ICL@2 IterChat & \textbf{0.5666} & \textbf{0.8875} & \textbf{0.6333} & \textbf{0.734} & 0.5284 & 0.7268 & 1.276 & 0.6194 &\textbf{0.1495}	&\textbf{0.5695}	&\textbf{2.2139} &\textbf{0.4378} \\
    \hline
    \multirow{4}{*}{full-parameter fine-tuning} & Llama-2-7B & multi-turn & 0.1667 & 0.5686 & 1.9 & 0.2488 & 0.4273 & 0.8981 & 0.9196 &0.6927 &0.4833	&0.8157	&0.9666 & 0.7112 \\
    
    & Llama-2-7B & IterChat & \textbf{0.3333} & \textbf{0.7002} & \textbf{1.1833} & \textbf{0.2840} & \textbf{0.7837} & \textbf{0.9363} & \textbf{0.3162} & \textbf{0.8213} &\textbf{0.8667}	&\textbf{0.9237}	&\textbf{0.1659} & \textbf{0.8659} \\
    \cline{2-15}
    & Llama-2-13B & multi-turn & 0.6666 & 0.8914 & 0.3833 & 0.4634 & 0.4704 & 0.9186 & 0.8436 &0.7151 &0.5	&0.8403	&0.85 & 0.9144\\
    
    & Llama-2-13B & IterChat & \textbf{0.8166} & \textbf{0.946} & \textbf{0.2166} & \textbf{0.5636} & \textbf{0.8181} & \textbf{0.9537} & \textbf{0.2542} & \textbf{0.8362} &\textbf{0.9667}	&\textbf{0.9871}	&\textbf{0.0937} & \textbf{0.9571} \\
    \hline 
    \end{tabular}
    }
    \caption{Preference Comparison between Multi-turn Dialogue and Iterchat}
    \vspace{-4mm}
    \label{tab:preference comparison}
\end{table*}

\section{Experiments}

\subsection{Experimental Setup}

\textbf{Datasets.} \textit{MultiWOZ 2.1} \cite{eric2019multiwoz} is a widely used, large-scale, multi-domain task-oriented dialogue (TOD) dataset with several revised iterations. For our study, we focus on the 'Hotel' domain of MultiWOZ due to its largest preference slots, which significantly increases task complexity. MultiDOGO \cite{peskov2019multi} another widely-used TOD dataset, is employed in our study. We specifically utilize its 'Software' domain subset for final evaluation. \textit{Foodie (IterChat)}, is a dataset designed for food preference extraction tasks, constructed using the IterChat data format. This dataset comprises 3,500 samples generated using GPT-4, with annotations validated by experienced data annotators to ensure quality. \textit{Foodie (multi-turn)} \footnote{Due to the company's security policies. We alternatively provide access to the IterChat version of the MultiWOZ-Hotel and MutiDOGO-Software datasets at https://github.com/walcheng/IterChat}, a dataset where transforms the History State in IterChat into multi-turn dialogues using GPT-4. These dialogues were then manually reviewed and corrected for logical consistency. Due to the complexity and effort on this process, we curated a subset of approximately 300 labeled dialogues for training and testing purposes.
\textbf{Evaluation Metrics.}
For easy comparison, we adopt the following metrics to evaluate the accuracy of preference extraction: (1) Exact Match (EM): Measures the percentage of predictions that exactly match the true labels;  (2) F1 scores: Harmonic mean of precision and recall, balancing both in one metric; (3) Filter Edit Distance (FED) \cite{li2024pctoolkit}: Counts the minimum changes needed to convert one string to another. (4) Bilingual Evaluation Understudy (BLEU) \cite{papineni2002bleu} measures the similarity between machine-generated text and reference translations. Since preference extraction tasks need to focus more on the correctness of preference extraction rather than the fluency of sentences, we set n-gram=1 to calculate the final BLEU score.

\textbf{Baseline models.}
We adopt popular open-source and close-source LLMs as baseline models for experiments, including GPT4  
 \cite{achiam2023gpt}, Llama-2 \cite{touvron2023Llama}, Qwen \cite{bai2023qwen} and Pangu \cite{ren2023pangu}. 

\textbf{Baseline Method.} \textit{PEARL} \cite{malik2024pearl} introduces a framework where large language models (LLMs) first summarize user preferences from dialogues, followed by a dynamic example retrieval module that stores and retrieves in-context learning (ICL) examples. \textit{NL2API}, another baseline, employs an LLM to take the demonstrations and preference slots in the prompt and then directly identifies the final preference label. Please refer to Appendix \ref{app: experimental settings} and Appendix \ref{app: prompt} for implementation details and the prompt.

\subsection{Multi-turn Dialogue vs. IterChat}
We evaluated the performance of multi-turn dialogues and IterChat under both few-shot prompting and full-parameter fine-tuning scenarios. In the few-shot setting, we used GPT-4 as the foundation model. 
For the MultiWOZ and MultiDOGO dataset, we employed 1,000 samples as the test set and 2 samples as demonstrations, while for the Foodie (IterChat/multi-turn) dataset, we used 60 samples as the test set and 2 samples as demonstrations. In the full-parameter fine-tuning scenario, we utilized 3,000 samples for training and 1,000 samples for testing on the MultiWOZ and MultiDOGO dataset, whereas, for the Foodie dataset (IterChat/multi-turn), we used 228 samples for training and 260 samples for testing. Based on the results in Table \ref{tab:preference comparison}, we observed the following: (1) In the few-shot prompting setting, the  data demonstration with IterChat data format does not demonstrate dominance across all datasets. This is because IterChat requires iteratively editing the preference set based on each user utterance, using operations such as adding, removing, or updating preference slot values. (2) However, in the full-parameter fine-tuning scenario, IterChat significantly outperformed the original multi-turn dialogues. This is because IterChat's data format, which includes only the historical state and the current dialogue, makes it easier for the model to learn preference transitions. In contrast, multi-turn dialogues involve tracking preferences across multiple turns, which often leads the model to converge to suboptimal solutions.

\begin{figure}[t] 
    \centering
    \vspace{-1mm}{\subfigure{\includegraphics[width=0.49\linewidth]{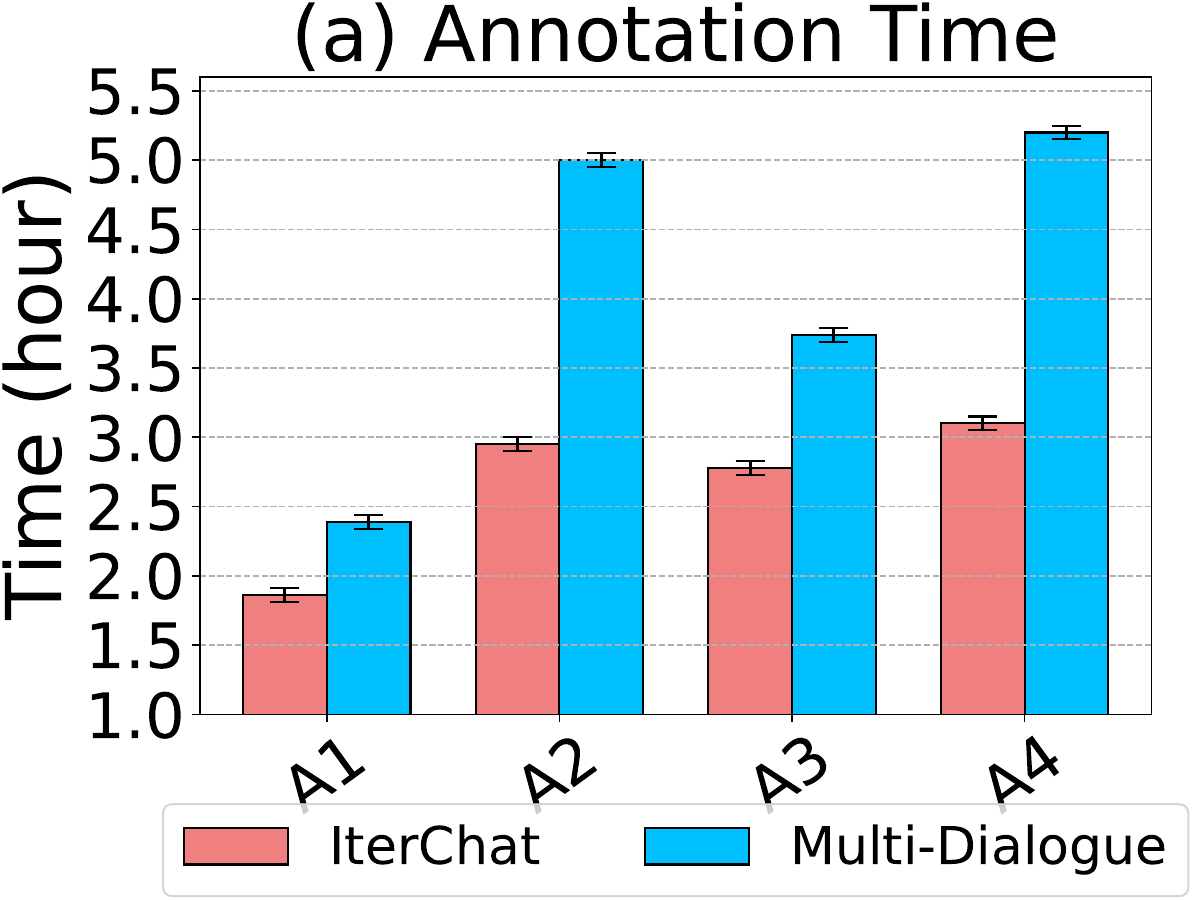}}}
	{\subfigure{\includegraphics[width=0.49\linewidth]{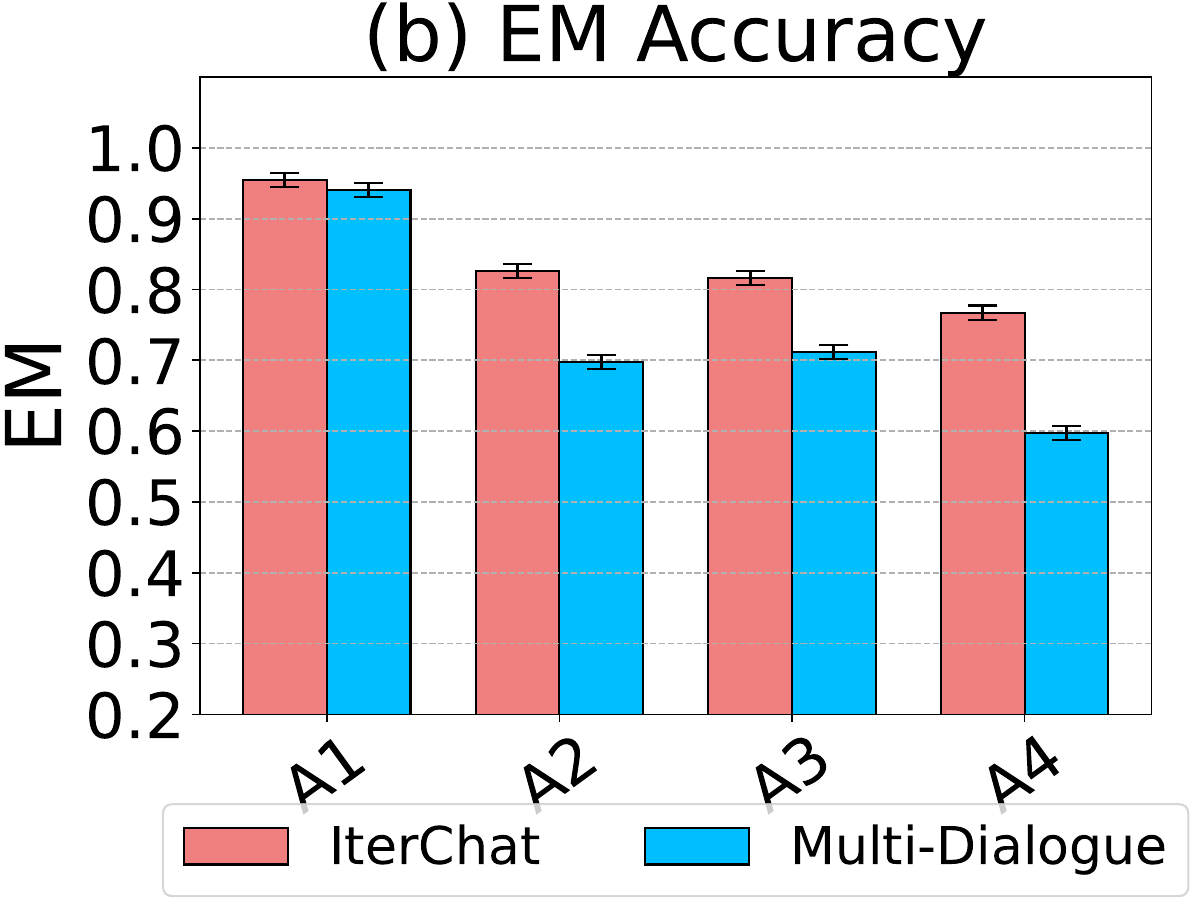}}}
    \vspace{-6mm}
    \caption{Annotation Efficiency and Accuracy.}
    \label{Figure:E2}
    \vspace{-5mm}
\end{figure}

\subsection{Annotation Efficiency and Accuracy}
To demonstrate the annotation efficiency of our proposed IterChat framework, we conducted an experiment comparing the efficiency and accuracy of human annotators when using IterChat versus traditional multi-turn dialogue context annotations. We employ one professional annotator (A1) and three temporary annotators (A2–A4) who joined the project. Each annotator independently labels 288 multi-turn dialogue instances as well as the data constructed by IterChat. The results of the experiment are shown in Figure \ref{Figure:E2} with the following insights: (1) \textbf{Annotation Time:} On average, the IterChat format significantly reduced the time spent on annotation, with an average time of 2.92 hours compared to 4.08 hours for the multi-turn dialogue format. This represents a 28.4\% reduction in annotation time, highlighting the efficiency gains of using the IterChat format. (2) \textbf{Annotation Accuracy (EM):} Despite the reduction in time, the IterChat format maintained or even improved accuracy in terms of Exact Match (EM) scores. The average EM accuracy for IterChat was 84.37\%, which is 11.95\% higher than the 73.42\% achieved with the multi-turn dialogue format.

\subsection{Generalization Ability}
To evaluate the generalization ability of our proposed IterChat, we conducted an experiment to assess how well various base models, after fine-tuning on the IterChat dataset, could generalize to unseen data. In this experiment, we fine-tuned three distinct base models of varying sizes on 3000 samples from the IterChat dataset. The selected models were Llama-2-13B and Qwen-32B, and PanGu-38B. Each of these models was fine-tuned on the IterChat dataset and then evaluated on a test set of size 200 to measure EM. The result is shown in Table \ref{Table:E3}.
\begin{table}[h!]
\centering
\resizebox{\linewidth}{!}{
\begin{tabular}{c|c|c|c}
\hline
\textbf{Model} & \textbf{Llama-2-13B} & \textbf{Qwen-32B} & \textbf{PanGu-38B} \\
\hline
\textbf{EM (pre-tuning)}    & 5\%            & 7\%           & 3.5\%            \\
\hline
\textbf{EM (post-tuning)}    & 84.0\%            & 86.5\%           & 83.0\%            \\
\hline
\end{tabular}
}
\caption{Model Performance (EM)}
\label{Table:E3}
\end{table}
\vspace{-1em}
After fine-tuning the IterChat dataset, the Llama-2-13B model achieved an EM accuracy of 84.0\%. It shows that even a general-purpose model with no specialized training in dialogue data can still benefit from the IterChat format. The Qwen-32B model achieved an EM accuracy of 86.5\%. This improvement over Llama-2-13B suggests that fine-tuning IterChat data helps improve its performance, possibly due to its inherent ability to handle context-rich information better than general-purpose models. On the other hand, PanGu-38B, despite its larger size, achieved the worst performance with an EM accuracy of 83.0\%. Despite its larger parameter count, PanGu-38B might not be as well-suited for the dialogue-based nature of the IterChat data.

The results of this experiment demonstrate the effectiveness of the IterChat format in enhancing the generalization capabilities of various base models. Across the three models tested, we observed consistent improvements in EM accuracy after fine-tuning on IterChat data.

\begin{figure} [h]
    \centering    \includegraphics[width=0.8\linewidth]{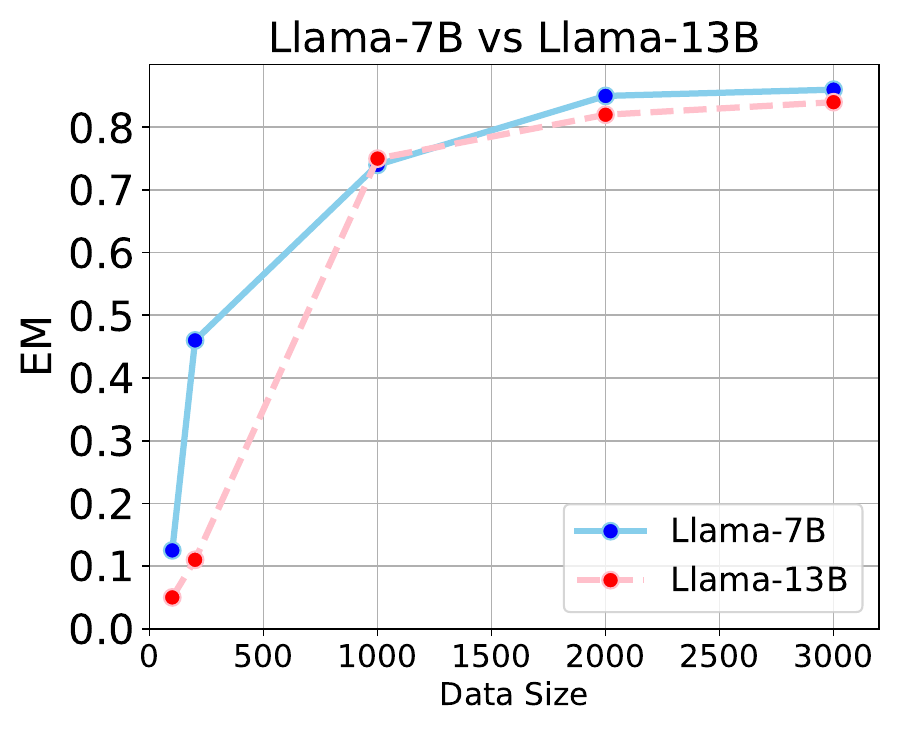}
    \vspace*{-4mm}
    \caption{Scaling of Llama-2 with different data size}
    \label{fig:E4}
    \vspace*{-4mm}
\end{figure}

\subsection{Training Scaling}

In this experiment, we demonstrate the scalability of our proposed IterChat dataset by evaluating how well fine-tuned large language models (LLMs) of different sizes perform as the amount of IterChat data increases. Specifically, we test the EM accuracy of two models: Llama-2-7B and Llama-2-13B after fine-tuning on IterChat datasets of different sizes, ranging from 100 to 3000 samples. After fine-tuning, both models were evaluated on a consistent set of 200 test samples to measure EM accuracy. The result is shown in Figure \ref{fig:E4}.

It can be observed that both models show a clear correlation between the size of the training data and the EM accuracy. Llama-2-7B, as a smaller model, benefits significantly from smaller datasets, achieving substantial performance improvements with even relatively few samples. On the other hand, Llama-2-13B requires a larger amount of data to fully leverage its larger capacity, with improvements becoming more noticeable after around 1000 samples. In conclusion, IterChat offers strong scalability, and both smaller and larger models benefit from increased training data.

\section{Conclusions}
In this work, we have presented IterChat, a novel framework for generating high-quality dialogue datasets that address the challenges of ``Annotating Disaster'' and ``Preference Oblivion'' in multi-turn dialogue preference extraction. By decomposing the task into more manageable one-turn preference extractions, IterChat enhances both the accuracy and efficiency of dialogue data annotation. The new format, which categorizes historical preferences separately from one-turn dialogues, reduces annotation errors and simplifies model training by alleviating the issues of error propagation across multiple turns. Our experiments show that fine-tuning or few-shot prompting with the IterChat format yields significantly improved performance in preference extraction tasks compared to the traditional multi-turn dialogue format. These findings underscore the potential of IterChat to both streamline the annotation process and improve the generalization capabilities of LLMs in dialogue systems.

\section*{Ethical Statement}
We have hired full-time data annotators and purchased the necessary work insurance for them. We strictly adhere to the regulation that the daily working hours shall not exceed 8 hours. Moreover, we offer salaries that are not lower than the market average.

\section*{Limitations}

\textbf{Dataset Construction and Generalization:} Although the paper claims that the IterChat format can enhance the generalization capabilities of various base models, the generalization ability evaluation is limited. The experiments mainly focus on a few datasets (such as the 'Hotel' domain of MultiWOZ and the Foodie dataset), and it is uncertain whether the results can be extended to other domains and more complex real-world scenarios. Also, the datasets used for evaluation may not fully cover the diversity of user preferences and dialogue situations in practice.

\textbf{Annotation and Data Generation:} The annotation process in IterChat still requires human effort, and although it reduces the annotation time and improves accuracy compared to multi-turn dialogues, it may still be resource-intensive for large-scale datasets. Additionally, the data generation process relies on GPT-4 to pre-define preference slots and sample values, which may introduce biases from GPT - 4 itself. Also, the assumption that multi-turn preference extraction can be decomposed into one-turn extraction processes might not hold true for all types of dialogues, especially those with highly complex and intertwined preference expressions.
\section*{Acknowledgments}

\bibliography{acl_latex}

\appendix

\section{Implementation Details.}
\label{app: experimental settings}
\textbf{Baseline Settings.} We evaluate the effectiveness of the IterChat data format compared to the original long-context format using two approaches: (i) few-shot prompting, each model directly processes the concatenated text of the task instruction and dialogue content (either IterChat or multi-turn dialogue) as the input prompt, generating the final set of preference labels. (ii) Full-parameter fine-tuning, we following the conventional one-dialogue-one-sample manner which the adopted baseline models are all causal LLMs.

\textbf{Parameter settings during full-parameter fine-tuninng.} We conducted full-parameter fine-tuning using distributed computing, employing a total of 8 distributed nodes. Each node was equipped with 72 CPU cores and 8 Huawei Ascend GPU, each with 64 GB of memory. During the fine-tuning phase, we set the batch size to 48 and trained the model for 5 epochs. For fine-tuning on small datasets (data size < 1,000), we used a learning rate of 4e-5, while for larger datasets (data size > 1,000), we set the learning rate to 5e-5. The Adam optimizer \cite{zhang2018improved} was used throughout all training processes.

\section{Prompts}
\label{app: prompt}
We provide the exact prompts utilized in our experiments on the MultiWoZ-H dataset, with similar prompts applied to the Foodie dataset.

\begin{figure*}
    \centering
    \includegraphics[width=1\linewidth]{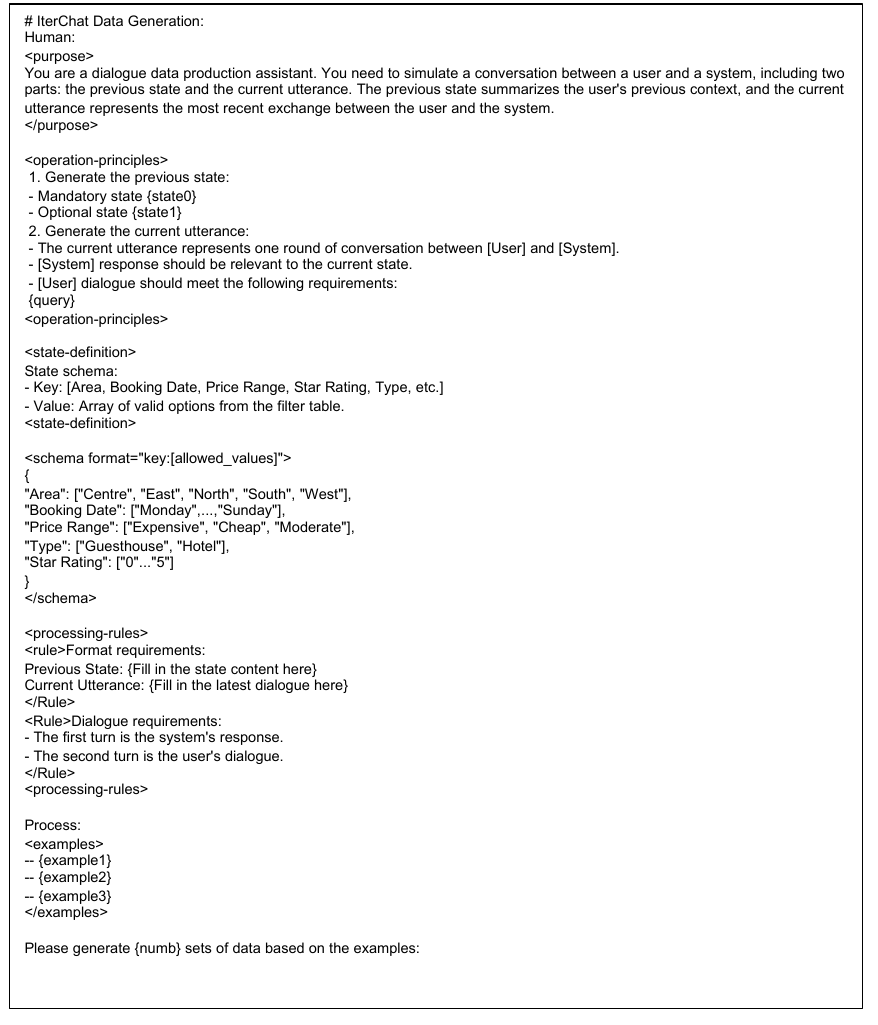}
    \caption{Prompt for new data format generation}
    \label{fig:enter-label}
\end{figure*}

\begin{figure*}
    \centering
    \includegraphics[width=1\linewidth]{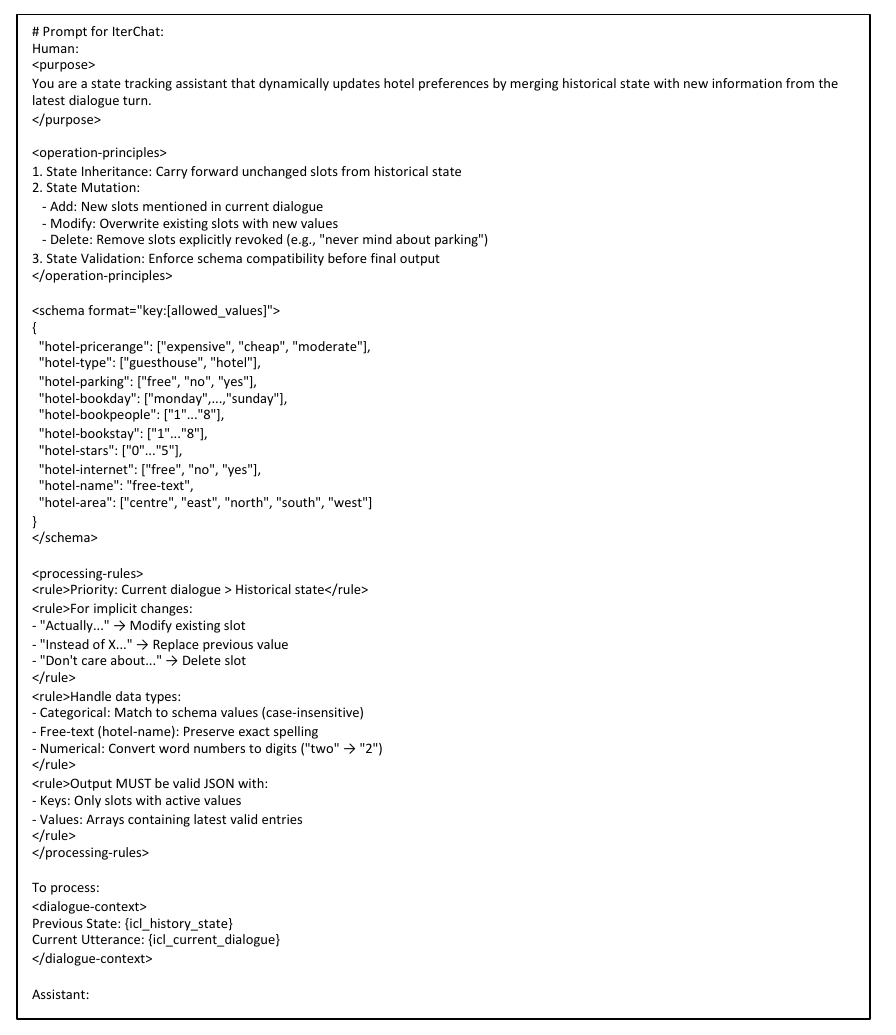}
    \caption{Prompt for IterChat during few-shot prompting}
    \label{fig: iterchat prompt}
\end{figure*}
\begin{figure*}
    \centering
    \includegraphics[width=1\linewidth]{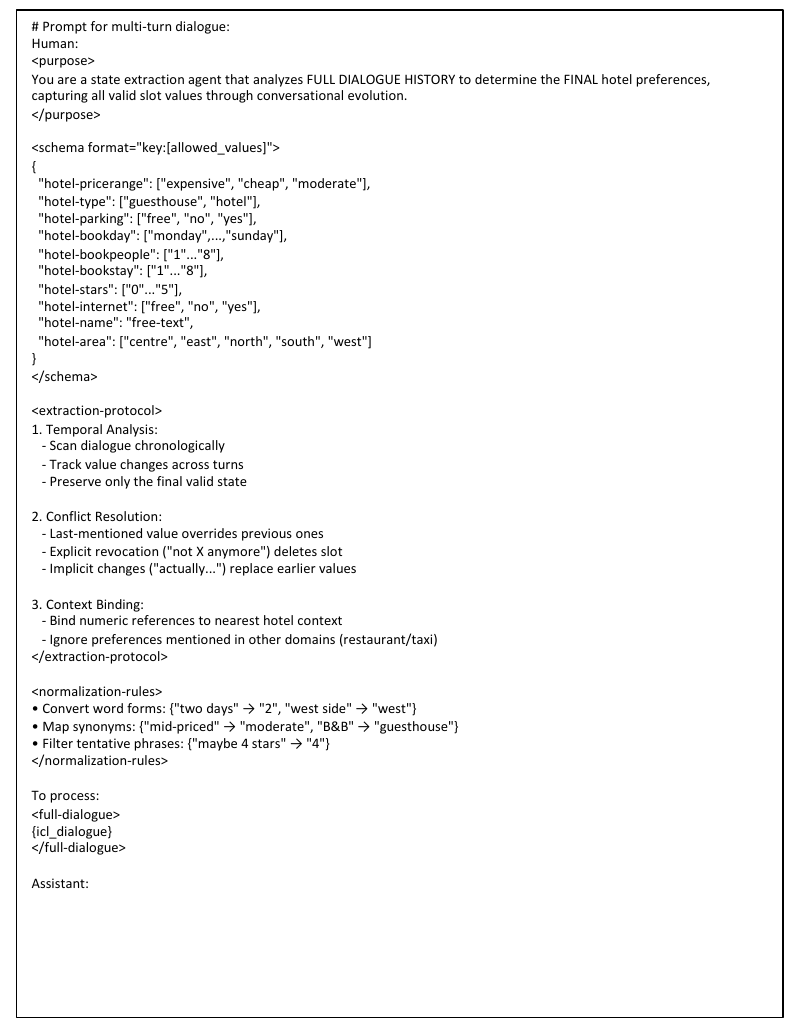}
    \caption{Prompt for multi-turn dialogue during few-shot prompting}
    \label{fig:multi-turn prompt}
\end{figure*}
\end{document}